\documentclass[letterpaper, 1+++++++0 pt, conference]{ieeeconf} 
\usepackage[T1]{fontenc}
\usepackage[latin9]{inputenc}
\usepackage{xcolor}
\usepackage{verbatim}
\usepackage{float}
\usepackage{amsmath}

\usepackage[T1]{fontenc}
\usepackage[latin9]{inputenc}
\usepackage{color}
\usepackage{verbatim}
\usepackage{float}
\usepackage{amsmath}

\usepackage{amsthm}
\usepackage{amssymb}
\usepackage{stmaryrd}
\usepackage{stackrel}
\usepackage{graphicx}
\usepackage{wasysym}
\usepackage{mathtools}
\usepackage{hyperref}
\usepackage{url}   
\usepackage{booktabs} 
\usepackage{nicefrac} 
\usepackage{authblk}
\usepackage{multirow}
\usepackage{xcolor}
\usepackage{microtype}
\usepackage{paralist}
\usepackage{subfigure}
\usepackage[ruled]{algorithm2e}
\usepackage{algpseudocode}

\makeatletter

\usepackage[normalem]{ulem} 

\theoremstyle{plain}

\theoremstyle{plain}

\theoremstyle{definition}
\newtheorem{defn}{\protect\definitionname}
\theoremstyle{definition}
\newtheorem{problem}{\protect\problemname}
\theoremstyle{plain}

\theoremstyle{definition}

\theoremstyle{remark}

\theoremstyle{plain}

\usepackage{cite}
\usepackage{algpseudocode}
\usepackage{setspace}
\IEEEoverridecommandlockouts

\usepackage{algpseudocode}
\usepackage{setspace}
\usepackage{color}
\usepackage{comment}
\usepackage{tikz}
\tikzset{>=latex}
\usetikzlibrary{fit,automata,positioning,angles,quotes}

\makeatother

\usepackage{babel}
\providecommand{\corollaryname}{Corollary}
\providecommand{\definitionname}{Definition}
\providecommand{\examplename}{Example}
\providecommand{\lemmaname}{Lemma}
\providecommand{\problemname}{Problem}
\providecommand{\remarkname}{Remark}
\providecommand{\theoremname}{Theorem}
\providecommand{\assumptionname}{Assumption}

\newcommand{\UntilOp}{\mathcal{U}}
\newcommand{\Eventually}{\diamondsuit}
\newcommand{\Always}{\square}
\newcommand{\Next}{\Circle}
\newcommand{\AP}{{AP}}

\usepackage{amsmath}               
{
	\theoremstyle{plain}
	\newtheorem{assumption}{Assumption}
	\theoremstyle{plain}
	
}

\SetSymbolFont{stmry}{bold}{U}{stmry}{m}{n}

\def\BibTeX{{\rm B\kern-.05em{\sc i\kern-.025em b}\kern-.08em
		T\kern-.1667em\lower.7ex\hbox{E}\kern-.125emX}}

\begin{document}

\title{\LARGE \bf
     Vision-Based Reactive Planning and Control of Quadruped Robots \\ in Unstructured Dynamic Environments
}

\author{Tangyu Qian, Zhangli Zhou, Shaocheng Wang, Zhijun Li, Chun-Yi Su, and Zhen Kan 

\thanks{This work was supported in part by the National Natural Science Foundation of China under Grant U2013601, 62173314 and CAAI-Huawei MindSpore Open Fund.}
\thanks{T. Qian, Z. Zhou, S. Wang, Z. Li, and Z. Kan (corresponding author) are with the Department of Automation, University of Science and Technology of China, Hefei, China. }
\thanks{C. Su is with the School of Automation and Guangdong Province Key Laboratory of Intelligent Decision and Cooperative Control, Guangdong University of Technology, Guangzhou, China}

}

\maketitle

\begin{abstract}
Quadruped robots have received increasing attention for the past few years. However, existing works primarily focus on static environments or assume the robot has full observations of the environment. This limits their practical applications since real-world environments are often dynamic and partially observable. To tackle these issues, vision-based reactive planning and control (V-RPC) is developed in this work. The V-RPC comprises two modules: offline pre-planning and online reactive planning. The pre-planning phase generates a reference trajectory over continuous workspace via sampling-based methods using prior environmental knowledge, given an LTL specification. The online reactive module dynamically adjusts the reference trajectory and control based on the robot's real-time visual perception to adapt to environmental changes.
\end{abstract}

\section{INTRODUCTION}

 Recent years have witnessed great advances in the locomotion capability of quadruped robots like the MIT Cheetah \cite{bledt2018cheetah} and ETH Anymal \cite{fankhauser2018anymal}. However, it's still rare to find one in daily scenarios, since the dynamic and unstructured environment poses significant challenges to existing methods \cite{leslie2022robots}. Hence, this work is motivated to develop a reactive planning and control strategy for quadruped robots with  temporal logic specifications and multi-modal perception to enable mission operation in unstructured dynamic environments.

Due to the capability of expressing complex robotic tasks beyond traditional point-to-point navigation \cite{li2018neural}, temporal logic-based motion planning has received significant attention ~\cite{Cai2021b, cai2023overcoming, chen2020temporal,zhou2022multiple}.  However, most of previous studies either focused on static environments \cite{chen2011formal} or assumed full observations of the environment by the robot \cite{Cai2023,li2021asymmetric,li2019human}. 
Recently, these methods were extended to dynamic environments \cite{Vasile2020,li2021safe,cai2020learning,li2022human,li2020human,li2023human}. Li et al. proposed a method that aims at finding the optimal solution within a limited time domain  \cite{Li2022}. The proposed algorithm is a sampling-based approach that searches for local optimal paths in dynamic environments to accomplish timed logic tasks\cite{otte2016rrtx}. Despite recent progress, the aforementioned works are built upon a key assumption that the goal is implicitly coupled with a specified location. This may significantly limit its applicability, as the goal can be time-varying or even infeasible, especially in a dynamic environment. A more nature idea is to formulate the robot's motion planning task based on the target objects. When considering the motion control of quadruped robots, various algorithms have been developed. For instance, Cheetah can reach a top speed of 3.7m/s using model predictive control and whole body control \cite{kim2019highly}.  A controlled backflip is achieved by Panther by representing the rotational dynamics using the rotation matrix \cite{ding2021representation}. Still, few prior works focus on solving tasks with complex temporal and logic constraints in unstructured and dynamic environments.

\begin{figure}[t]
    \centering  \includegraphics[width=0.48\textwidth]{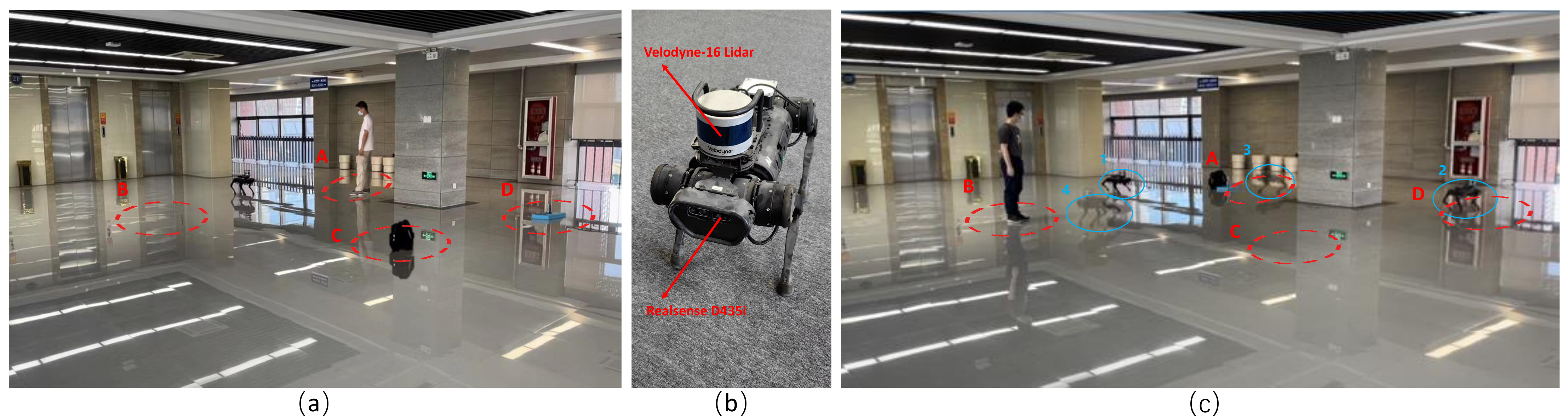}
    \caption{ (a) The experiment environment. (b) The Unitree A1 quadruped robot. (c) The visualized robot trajectory.}
    \label{real_exp}
 \end{figure}

To this end, we consider a robot integrated with vision performing high-level linear temporal logic (LTL) tasks that encode position constraints to the target object, as shown in Fig. \ref{real_exp}. To cope with unstructured dynamic environments, we developed a vision-based reactive planning and control (V-RPC) framework consisting of two layers: offline pre-planning and vision-based online reactive motion planning. The main contributions can be summarized as follows. We propose a novel task description that decouples the target object from the environment to enable efficient motion planning in unstructured dynamic environments. A real-time local scene perception algorithm is then developed, which can effectively detect and localize dynamic elements of environments and update robot's knowledge of the environment. Finally, targets of the motion control algorithm are determined by the scene perception module, which helps to adapt the robot to complex environments. Extensive numerical simulations and physical environments using the Unitree A1 quadruped robot are carried out to demonstrate the effectiveness of V-RPC. Fig. \ref{real_exp} shows the robot hardware and  the experiment setup .


\section{Backgound}

\subsection{Preliminaries}
	
Considering a robot operating in an unstructured dynamic  environment $Env$, the interaction between the robot and $Env$ can be captured by a weighted transition system (WTS).
	
\begin{defn}
    \label{def:WTS}
    A WTS in $Env$ is a tuple $\mathcal{T}=(X, x_{0}, \rightarrow_{\mathcal{T}}, \AP, L_{X}, C_{\mathcal{T}})$, where $X$ is the geometric space of $Env$; $x_{0}$ is the initial state of robot; $\rightarrow_{\mathcal{T}} \subseteq X\times X$ is the geometric transition relation s.t. $\left(x,x'\right)\in\rightarrow_{\mathcal{T}}$ if $dist(x, x')\leq\eta$, where $\eta$ is a predefined maximum step size, and the transition from $x$ to $x'$ is collision-free; $\AP$ is the set of atomic propositions indicating the labels of regions; $L_X: X\rightarrow\AP$ is the time-varying labeling function that returns an atomic proposition satisfied at the current location $x$; and $C_{\mathcal{T}}: (\rightarrow_{\mathcal{T}})\rightarrow\mathbb{R}^{+}$ is the geometric Euclidean distance, i.e., $C_{\mathcal{T}}(x,x')=dist(x,x'),\forall (x,x')\in\rightarrow_{\mathcal{T}}$.
\end{defn}


Due to the rich expressivity, LTL is used throughout this work to describe the high-level robot missions.


\begin{defn}
\cite{baier2008} (Linear temporal logic) LTL is a formal language over a set of atomic propositions $\AP$ and combinations of Boolean and temporal operators. The syntax of LTL is defined as:
\begin{equation*}
				\phi   :=  \text{true} \, | \, ap \, | \, \phi_1 \land \phi_2 \, | \, \lnot \phi_1 | \Next\phi \, | \, \phi_1 \UntilOp \phi_2\:, 
			\end{equation*}
			where $ap\in\AP$ is an atomic proposition, \emph{true}, \emph{negation} $\lnot$, and \emph{conjunction} $\land$ are propositional logic operators, and \emph{next} $\Next$ and \emph{until} $\UntilOp$ are temporal operators. Based on that, other propositional logic operators such as \emph{false}, \emph{disjunction} $\lor$, \emph{implication} $\rightarrow$, and temporal operators such as \emph{always} $\Always$ and \emph{eventually} $\Eventually$, can be defined.
		\end{defn}

Any LTL formula can be converted to a Non-deterministic B\"uchi Automata (NBA).

\begin{defn}
     An NBA over $2^{\AP}$ is a tuple $\mathcal{B}=(Q, Q_{0}, \Sigma, \rightarrow_{\mathcal{B}}, Q_{F})$, where $Q$ is the set of states, $Q_{0}\subseteq Q$ is the set of initial states, $\Sigma=2^{\AP}$ is the finite alphabet, $\rightarrow_{\mathcal{B}}\subseteq Q\times\Sigma\times Q$ is the transition relation, and $Q_{F}\subseteq Q$ is the set of accepting states.
\end{defn}
		


\subsection{Gait Representation}

The gait of a quadruped robot refers to the relative temporal relationship between stance and swing phase of 4 dependent legs. The gait can be divided into different categories based on the duty factor and relative phase. The duty factor $\rho$ is the quotient of stance duration $T_{stance}$ by the full gait cycle duration $T_{gait}$, where $T_{gait}$ = $T_{stance}$ + $T_{swing}$ with $T_{swing}$ representing the swing duration.
The relative phase $\varphi$ describes the radio of the time difference between legs in the whole gait cycle. If a leg is chosen as reference, its relative phase is set to 0. To define the gaits, we use the array [$\varphi_{LF}$, $\varphi_{LH}$, $\varphi_{RH}$, $\varphi_{RF}$] to represent the relative phases. Further details about the gait control are referred to Sec \ref{gait control}.

\section{Problem Formulation}
\label{Problem}

Consider a quadruped robot operating in an unstructured dynamic environment $Env$ and equiped with an RGB-D camera. Initially, the robot only has a preliminary map $\mathcal{M}$ of the environment.
The problem to be solved is formally stated as follows.

\begin{problem}
\label{Prob1} Given a constrained task $\phi_c$ and a quadruped robot with the initial position  $x_0$, the goal is to design a vision-based reactive motion planning strategy such that $\phi_c$ is ensured to be completed in the dynamic environment $Env$.
\end{problem}

Due to the consideration of unstructured dynamic environments, there are two main challenges in solving Problem \ref{Prob1}: the \emph{inefficiency challenges} and the \emph{ incompleteness challenges}. The inefficiency challenges represent the emergence of a new planning more efficient in accomplishing the task. And the incompleteness challenges indicate infeasible subtasks  and thus a modified planning is needed. 

\begin{assumption}
\label{assum1}
The constrained task $\phi_c$ can be accomplished regardless how the environment $Env$ changes.
\end{assumption}

Assumption \ref{assum1} is mild and reasonable. Otherwise, the task $\phi_c$ cannot be completed no matter how the motion planning is designed and there is no need for further discussion. 


\section{Method}

\begin{figure}[ht]
\centering
\includegraphics[width=3.2in]{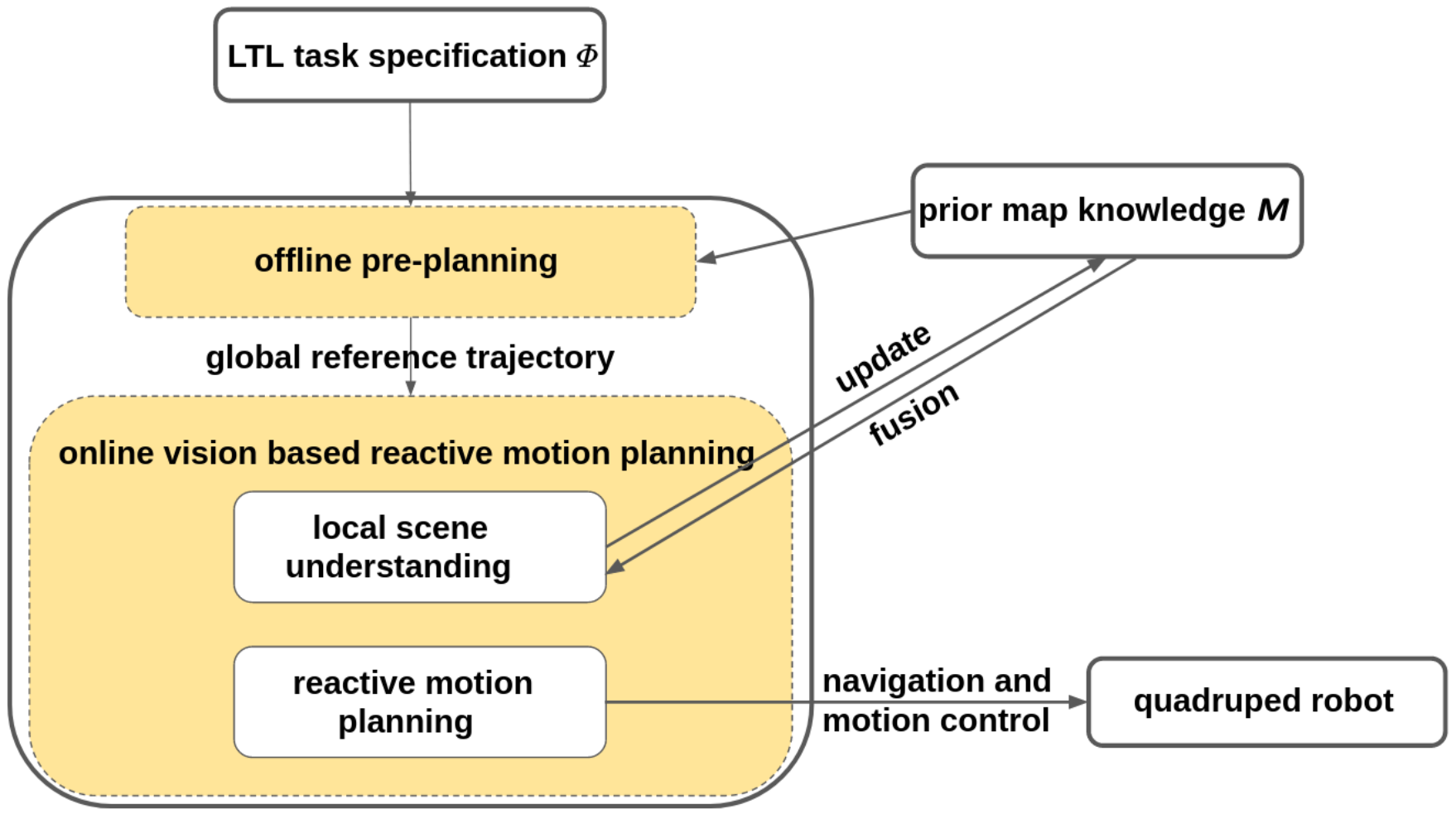}
\caption{Framework of V-RPC: vision-based reactive planning and control.}
\label{flow_chart}
\end{figure}

The developed V-RPC consists of a two-layer planning, as shown in Fig. \ref{flow_chart}. Sec. ~\ref{off_pre} introduces a pre-planning method as a top layer, to generate global trajectories for LTL satisfaction. 
By taking offline planning as reference trajectories, as the bottom layer, reactive local planning and control based on sensor fusions is developed in Sec. ~\ref{online_reactive}.

\subsection{Offline Motion Planning \label{off_pre}}
In this section, the robot's a prior knowledge of the environment $\mathcal{M}$ and LTL task specification $\phi$ are used to generate a pre-planning. 
Due to the time-varying labeling functions in the dynamic environment, the target objects are specified using atomic propositions with constraints, such as $\phi_c$. Note that $\phi_c$ can be converted to $\phi$

 \begin{defn}
	\label{def:PBA}
	Given the WTS $\mathcal{T}$ and the NBA $\mathcal{B}$, the product B\"uchi automaton (PBA) is a tuple $P=\mathcal{T}\times\mathcal{B}=(Q_P, Q^{0}_{P}, \rightarrow_{P}, Q^{F}_{P}, C_{P}, L_{P})$, where $Q_{P}=X\times Q$ is the set of infinite product states, $Q^{0}_{P}=x_{0}\times Q_{0}$ is the set of initial states; $\rightarrow_{P}\subseteq Q_{P}\times 2^{\AP}\times Q_{P}$ is the transition relation defined by the rule: $\frac{x\rightarrow_{\mathcal{T}}x'\land\text{ } q\overset{L_X(x)}{\rightarrow_{\mathcal{B}}}q'}{q_{P}=(x,q)\rightarrow_{P} q_{P}'=(x',q')}$, where $q_{P}\rightarrow_{P} q_{p}'$ denotes the transition $(q_{P},q_{P}')\in\rightarrow_{P}$, $Q^{F}_{P}=X\times Q_{F}$ is the set of accepting states, $C_{P}\colon (\rightarrow_{P})\rightarrow\mathbb{R}^{+}$ is the cost function defined as the cost in the geometric space, e.g., $C_{P}(q_{p}=(x,q),q_{p}'=(x',q'))=C_{\mathcal{T}}(x,x'), \forall (q_{P},q_{P}')\in\rightarrow_{P}$, and $L_{P}\colon Q_{P}\rightarrow\AP$ is the labelling function s.t. $L_P(q_{P})=L_X(x), \forall q_{P}=(x,q)$.
\end{defn}

A valid trace $\tau_{P}=q^{0}_{P}q^{1}_{P}q^{2}_{P}\ldots$ of a PBA is called accepting, if it visits $Q^{F}_{P}$ infinitely often. The corresponding accepting word $\tau_{o}=o_{0}o_{1}o_{2}\ldots, \forall o_{i}=L_{P}(q^{i}_{P})$ satisfies the corresponding LTL formula $\phi$. Let $\tau_{F}$ denote an accepting trace and $proj|_{X}: Q_{P}\rightarrow X$ a function that projects the product state space into the workspace, i.e., $proj|_{X}(q_{p})=x, \forall q_{P}=(x,q)$. Using the projection, we can extract a trajectory $\tau_{\mathcal{T}}=proj|_{X}(\tau_{F})$ that satisfies the LTL formula.
More details are presented in~\cite{baier2008}. Therefore, the goal of pre-planning is to find an accepting path $\tau_{P}$ of PBA, with minimum accumulative geometric cost $C_{P}$.

If the state space of WTS is continuous, it's impossible to explictly construct a PBA. Hence, the sampling-based method TL-RRT* 
from ~\cite{luo2021abstraction} is employed to track PBA on-the-fly, which can generate the feasible optimal path for LTL satisfaction. The resulting trajectory is a lasso-type sequence in the form of prefix-suffix structure, i.e. $\tau_{F}=\tau^{pre}_{P}[\tau^{suf}_{P}]^{\omega}$, where the prefix part $\tau^{pre}_{P}=q^{0}_{P}q^{1}_{P}\ldots q^{K}_{P}$ is only executed once, and the suffix part $\tau^{suf}_{P}=q^{K}_{P}q^{K+1}_{P}\ldots q^{K+M}_{P}$ with $q^{K}_{P}=q^{K+M}_{P}$ is executed infinitely often.
Following prior work~\cite{cai2023overcoming}, we can decompose the optimal path  $\tau_{F}=\tau^{pre}_{P}[\tau^{suf}_{P}]^{\omega}$ based on automaton components into a sequence of goal-reaching trajectories i.e., $\tau_{F}=\tau_{0}\tau_{1}\ldots\tau_{K}[\tau_{K+1}\ldots \tau_{K+l}]^{\omega}$. Each $\tau_{i}$ can be presented as an optimal solution of reachability navigation expressed as a simple LTL formula $\phi_{i, F}=\Always\lnot\mathcal{O}\land\phi_{g_{i}}$, $\mathcal{O}$ represent obstacles.  We refer readers for more details about the decomposition procedure in~\cite{cai2023overcoming}. 

In the following section, we take each $\tau_{i}$ as a reference path and synthesize a vision-based reactive motion planning algorithm through RGB-camera to satisfy each $\phi_{i, F}$ in dynamic  environments.

\subsection{Online Reactive Motion Planning \label{online_reactive}}

\subsubsection{Multimodal Sensing\label{local_under}}


We implement vision sensors to enable the robot to continuously update its prior knowledge. To extract the high-level representation of the scene from the color image $r$ of the RGB-D camera, the darknet from \cite{bochkovskiy2020yolov4} is used to detect the classes of task-relevant objects in the field of view of camera. All 2D bounding boxes $B$ related to the task are then obtained. Point cloud is segmented and clustered using the density-based spatial clustering of applications with noise ~\cite{ester1996density} so that the centroids and the respective convex hull dimensions can be obtained.

 \subsubsection{Scene Understanding} \label{scene understanding}
 
Human motion plays an essential role in the environment. Therefore,we came up with a way that utilized a cylinder with a radius of 0.2$m$ and 2$m$ high to determine the best position based on the occupancy of the point cloud. We defined the occupancy value in terms of the fraction $frac$, where $frac = \frac{n}{100 \times N} \%$, where $N$ refers to the number of point clouds belonging to humans, and $n$ represents the number of point clouds belonging to both humans and cylinders.
Generally, the more the point cloud occupies the cylinder's volume after clustering, the more accurate the prediction is. 
 
 Furthermore, we devise a method to determine whether the drifting of the target object's predicted position results from the target object's movement or the jitter induced by the robot movement. To address this issue, we set a threshold value to the movement speed of dynamic objects in the environment. If the drift exceeds this threshold, the object is considered dynamic and, conversely, a drift caused by robot bumps and sensor errors. 
 The aforementioned solutions can significantly enhance the accuracy of predicting target objects and differentiate drifting caused by measurement errors or dynamic objects. 





 


\begin{figure}[t]
\centering
\includegraphics[ width=3.2in]{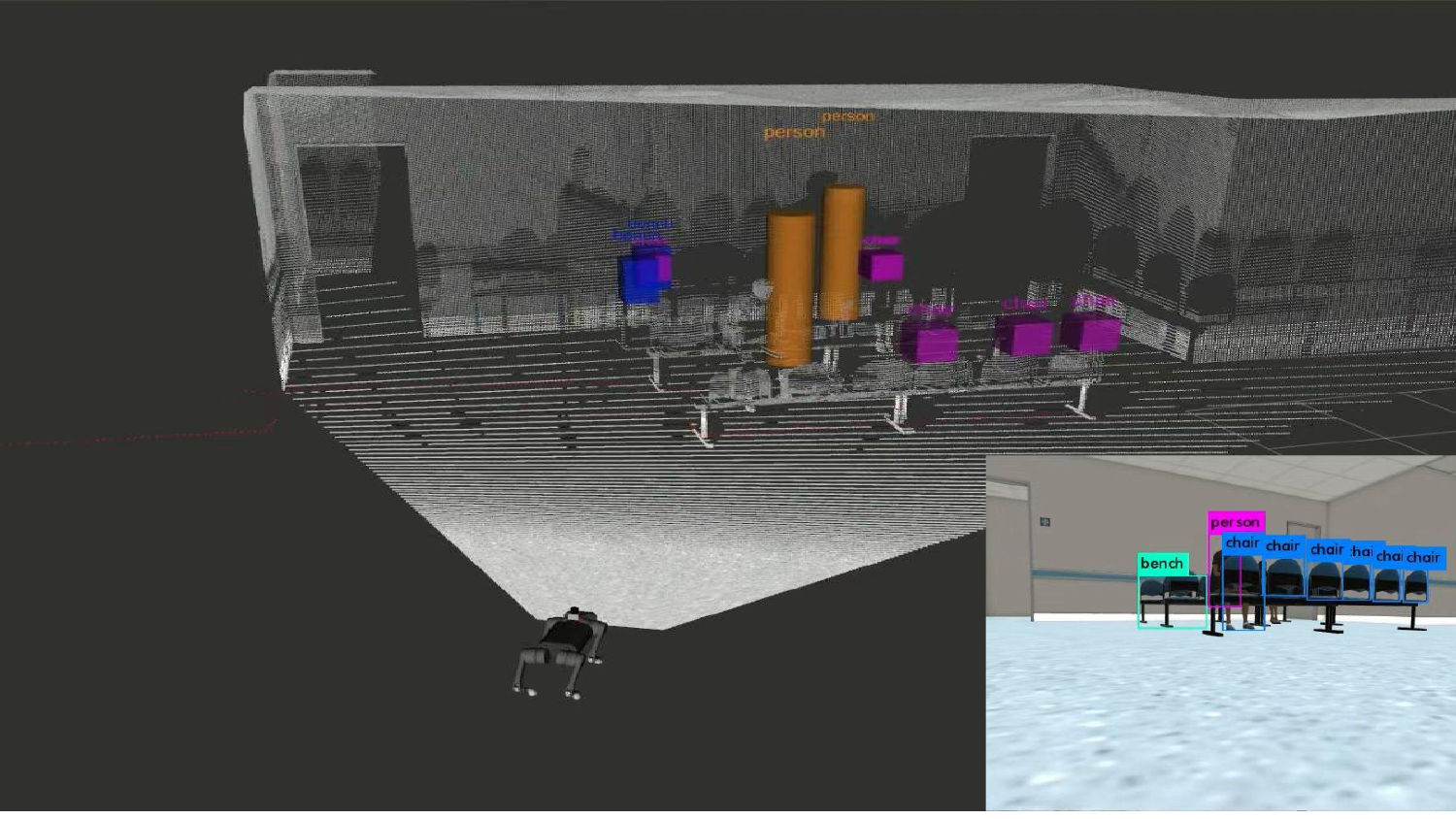}
\caption{Rviz visualization: The robot uses local visual information to update its own knowledge of the environment}
\label{vision_location}
\end{figure}

\subsubsection{Reactive Motion Planning and Control \label{Vision_Reactive}}

\paragraph{Planning}

Given an LTL task specification $\phi$ corresponding to the constrained task $\phi_c$, the labeling function $L_X^B$ is used to record the position constraint of the target objects. We can generate offline trajectories $\pi$ that satisfies $\phi$ according to Section \ref{off_pre}. Then, the trajectory $\pi$ can be projected onto $\mathcal{T}$ to obtain a list of target object's location $locate$ that need to be accessed, along with the corresponding list of position constraints $bind$.
In the following, we show how to adjust offline reference trajectories for dynamic environments using online vision-based reactive motion planning module.

Two mechanisms i.e., the greedy mechanism and the gate mechanism are introduced to deal with the \emph{inefficiency challenges} and \emph{incompleteness challenges} in Section \ref{Problem}. 

The greedy mechanism deals with two main types of situations. In the first situation, when the robot intended to go to $locate[i]$ to execute the subtask $x_i$, it may find a new location $x^{\star}$ closer to the current state $x(t)$ while satisfying position constraints $bind$. Mathematically, $dist(locate[i],x(t)) > dist(x^{\star}, x(t)) \land x^{\star} \in bind[i]$. At this point, rather than continue with the previous goal $locate[i]$, $x^{\star}$ will be selected instead as the new goal. Another situation is that the current subtask cannot be completed and the robot's prior knowledge does not provide valid information. In this case, the robot goes to the vicinity of the next subtask to find the current target object, and if it finds a solution to complete the current subtask on the way, it can execute it.

The gate mechanism: When the intended location of the subtask $locate[i]$ is reached, the subtask may not be completed as expected due to the change of environment, and it is necessary to find a new target point $N(x_i)$ based on $\mathcal{M}$. This point is the closest point to the robot that can satisfy the requirement $bind[i]$. Combining the above steps, we can obtain the optimal trajectory $\tau*$ that satisfies the task $\phi_c$ on $\mathcal{T}$. It is easy to see that this reaction process is not on the automaton, so there is no need to reconstruct the automaton or re-search such a time-consuming and labor-intensive operation when a dynamic response occurs in the environment. Thanks mainly to the target object rather than the position in the environment used for the LTL task description, the robot is able to choose how to carry out reactive planning according to the task autonomously. 


 \begin{figure}[t]
	\centering 	\includegraphics[width=0.3\textheight]{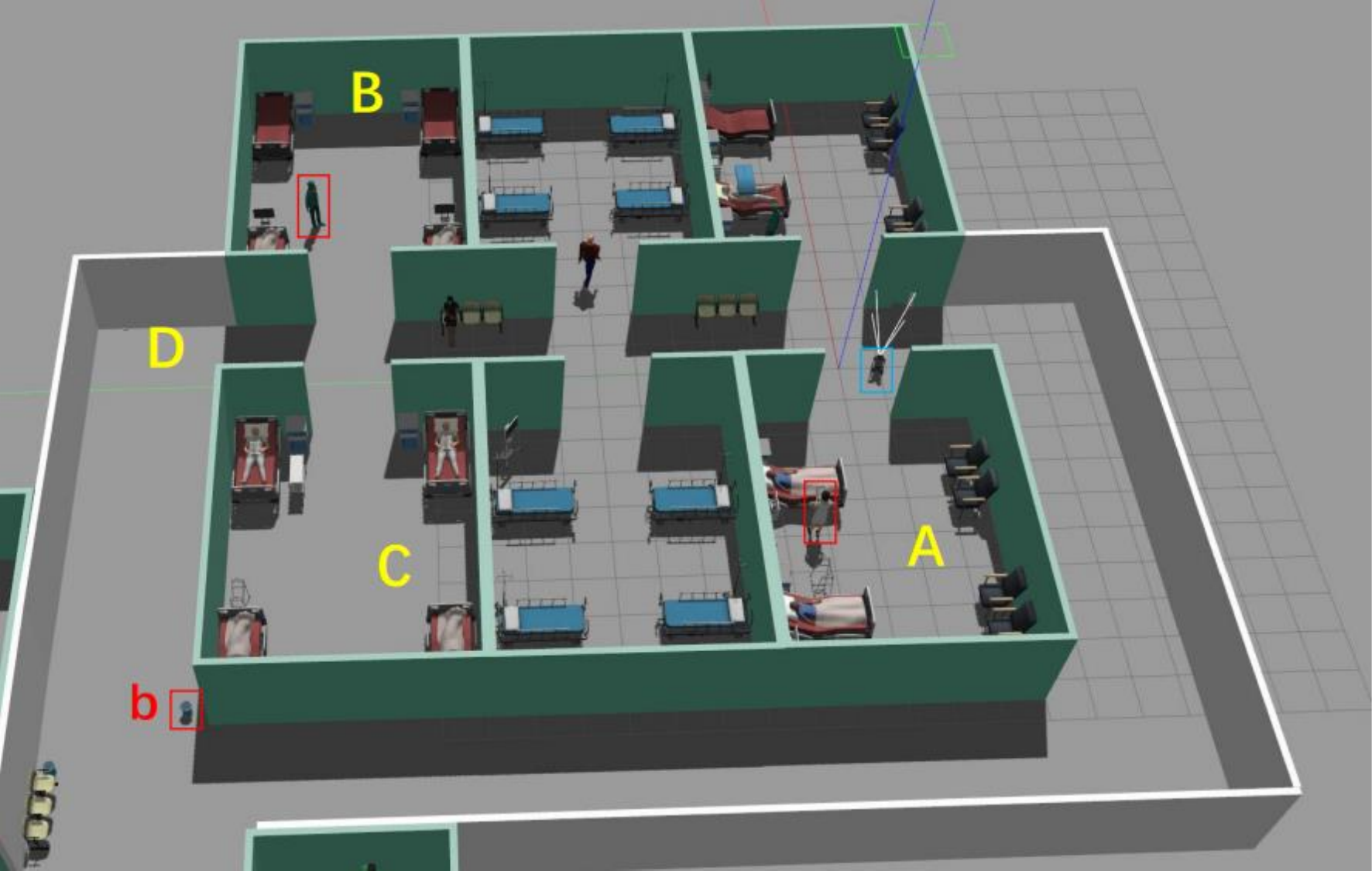}
	\caption{A quadruped robot is asked to perform the task $\phi_c$ in a hospital environment. The robot needs go to room $A$ to get the prescription and delivers it to the patient in room $B$. After taking the medicine, the medicine box should be dropped in the trash bin $b$ in the platform $D$ and the medicine should be returned to the nurse in room $A$. }
	\label{hos2}
\end{figure}

\paragraph{Control} \label{gait control}

 In Sec. \ref{online_reactive}, we introduced how to detect and classify different objects. Afterwards, the resulting information is utilized for reactive navigation planning. In this paragraph, prior map knowledge and object detection information are fused to implement scene understanding-guided reactive motion control.
 
The gaits can be classified as dynamic ones or static gaits. Depending on the scene understanding, the quadruped robot switches between this two kinds of gaits.
The static gait is defined as having at least three legs in contact with the ground at the same moment. Therefore the center of pressure (CoP) of torso always projects within the support polygon generated by the stance legs, allowing for better stability and adaptability when encountering unstructured terrains. Having at most two legs in contact with the ground at the same time, the dynamic gait gives rise to faster velocity and improved energy efficiency in despite of the underactuation. 

In this work, we consider dynamic trot gait and static walk gait. In the trot gait, the two diagonal legs move at the same time. Due to the diagonal symmetry, relative phase $\varphi$ requires to be [0, 0.5, 0, 0.5]. When duty factor exceeds 0.5, two full stance phases appear, leading to better stability and slower speed, and vice versa. That is, the smaller the duty cycle, the faster the motion speed. Since a full flight phase might induce an impact to the torso, we choose $\rho$ = 0.5 with $T_{gait}$ = 0.6s.  for the trade-off between motion speed and stability.

 \begin{figure}[t]
	\centering 	\includegraphics[width=0.35\textheight]{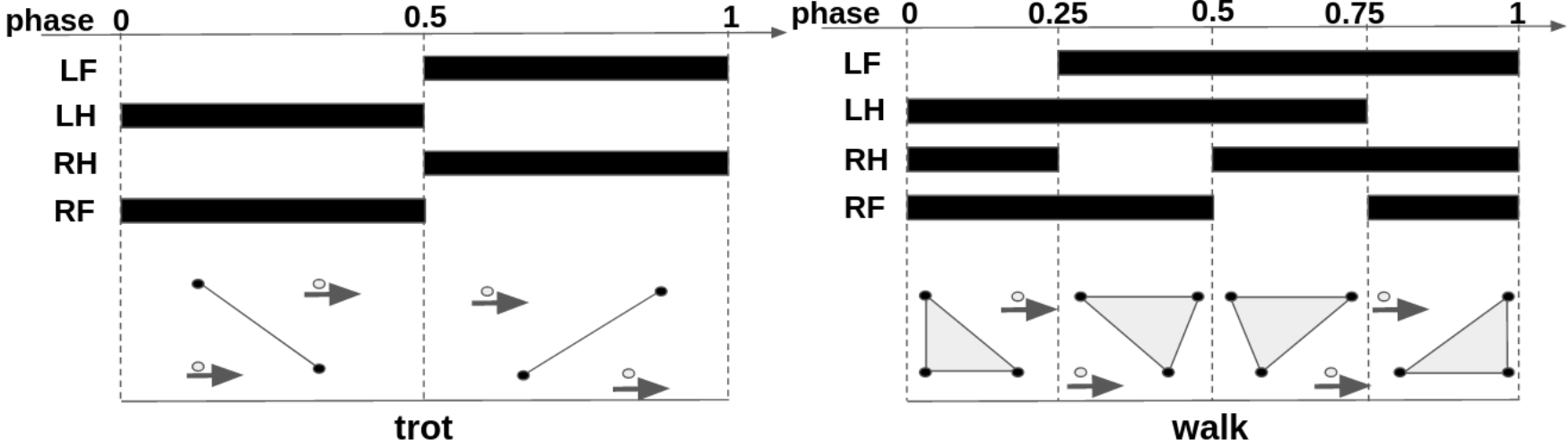}
	\caption{The grey triangle represents the support polygon. The black blocks represent the stance action, while the white represents the swing action. The solid dot represents a stance foot. The hollow dot represents a swing foot and the arrows indicate the swing direction.}
	\label{twoway}
\end{figure}


Four legs swing in turn in a certain order when executing the walk gait. By default static gait, the duty factor should be no less than 0.75. Hence, We set duty factor as $\rho$  = 0.75 for higher motion frequency. Meanwhile, the relative phase is set as [0, 0.75, 0.25, 0.5] for the sake of gait alignment. Additionally, $T_{gait}$ is set to 1.2s.

Since different stance poses in one gait can be treated as finite predefined modes of the switched system, following \cite{liao2022walking} and \cite{sleiman2021unified}, the gait control of a quadruped robot is formulated as a nonlinear model predictive control  (NMPC) problem
\begin{subequations}
\begin{align}
& \mathop{\min}\limits_{u_k} \sum\limits _{k=0}^{N-1} l_k (x_k,u_k) \label{Za}\\
& s.t. \quad x_0 = x(0) \label{Zb}\\
& \quad \quad \dot{x} = f(x(t), u(t), t) \label{Zc}\\
& \quad \quad g_k(x_k,u_k)=0, \quad k=0,...,N \label{Zd}\\
& \quad \quad h_k(x_k,u_k) \ge 0, \quad k=0,...,N \label{Ze}
\end{align}
\end{subequations}
where $x=\left[h_{com}^{T},q_{b}^{T},q_{j}^{T}\right]^{T}\in\mathbb{R}^{24}$
denotes the system states and $u=\left[f_{c}^{T},v_{j}^{T}\right]^{T}\in\mathbb{R}^{24}$
denotes the inputs, where
$h_{com}$ is the collection of normalized centrodial momentum, $q_{b}$ is the generalized coordinate of base, $q_j$ and $v_j$ are the joint positions and velocities, and $f_c$ consists of contact forces at four contact points. In the NMPC, (1a) is the tracking cost to follow a given trajectory in $N$ horizons, (1b) is the initial state given by the state estimator, (1c) represents the centrodial dynamics, (1d) and (1e) are the equality and inequality constraints for specific robot tasks. The NMPC solver \cite{OCS2} computes the optimal trajectory of the switched system. The optimal trajectory $x^*$ and $u^*$ are then tracked by a whole body controller (WBC) in a hierarchical optimization fashion \cite{bellicoso2016perception} to get the feed-forword torque $\tau_{ff}$. Finally, the motor torque is computed by a PID controller.

\section{SIMULATION AND EXPERIMENT}
In this section, we experimentally validate the effectiveness of vision-based reactive planning and control (V-RPC) framework in simulation and real-world environments.

\subsection{Simulation}

	\label{Simulation}
 
\subsubsection{Reactive Navigation}  
\begin{figure*}[t]
	\centering
    \includegraphics[width=\textwidth]{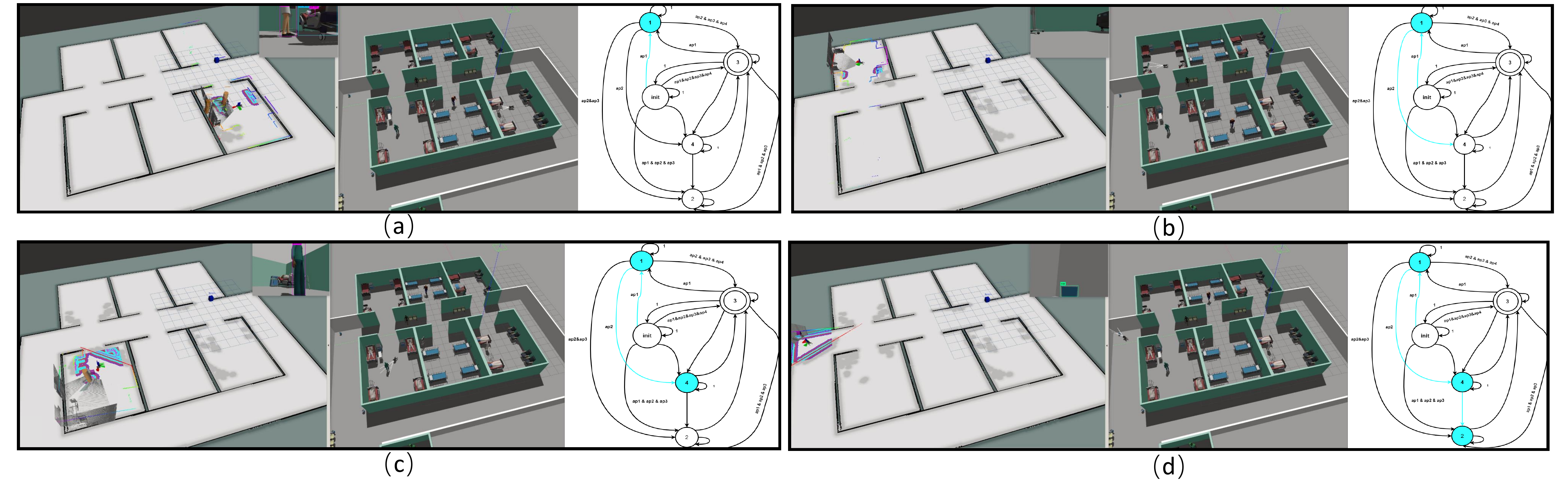}
	\caption{ Visualization of the task $\phi_c$. (a) The successful completion of ${nurse}^{room_A}$, (b) The gate mechanism declines the state transition in NBA $\mathcal{B}$. (c) The successful completion of $doctor^{Room_B \vee Room_C}$.  (d) The successful completion of $can$.}
	\label{sim}
\end{figure*}

The constrained task is specified as $\phi_c = \boxempty\diamondsuit nurse^{Room_A}\wedge\boxempty\diamondsuit doctor^{Room_B \vee Room_C}\wedge\boxempty\diamondsuit can\wedge\boxempty\diamondsuit nurse^{Room_A}$, where 
$nurse$ represents nurses, $doctor$ represents doctors, and $can$ represents trash cans. The superscript represents position constraints and empty means no constraints. The task $\phi_c$ requires the quadruped robot to go to room $A$ to find the nurse to consult the patient's condition, and then go to room $B$ or room $C$ to find a doctor and get medicine. Before bringing the medicine to the nurse in room $A$, it needs to throw the box of medicine into a random trash can. In Fig .\ref{hos2}, the target objects where the robot performs its task are marked with red boxes. Given the current environment, a feasible solution for $\phi_c$ is to go from room $A$ to room $B$, then goes to the trash can $b$ in the corridor $D$ to throw the trash, and finally returns to $A$. If $Env$ is dynamic and unstructured, as shown in Fig. \ref{hos2}, the doctor may walk around between room $B$ and room $C$ and there might exist other trash cans $a$ in the corridor $D$. Hence, the robot may not find the doctor when arriving at room $B$. This will trigger the gate mechanism. Specifically, the current subtask is considered as not completed yet, even if the robot arrives at room $B$ as pre-planned. The gate mechanism declines the state transition in NBA $\mathcal{B}$, as shown in Fig. \ref{sim}(b). Since the subtask of finding a doctor to take the medicine could not be completed as desired, the robot needs to re-plan its motion by going to room $C$ to look for the doctor. When reaching room $C$ and finding the doctor, the gate mechanism confirms the completion of the current subtask and the task then $\phi_c$ proceeds, as shown in the state transition in Fig. \ref{sim}(c)). In Fig. \ref{sim}(d), the robot finds another trash can $a$ on its way to the trash can $b$. Since $a$ is closer than $b$ to its current location, the greedy mechanism kicks in allowing the robot to choose the trash can $a$ instead to throw the medicine package. After the medicine package has been trashed at $a$, the gate mechanism confirms the completion of task $can$ and proceeds to the last subtask. More details are referred to the experiment video\footnote{\url{https://youtu.be/CbEbulAc0O4}}.

		


 \begin{figure}[t]
	\centering
    \includegraphics[width=0.4\textwidth]{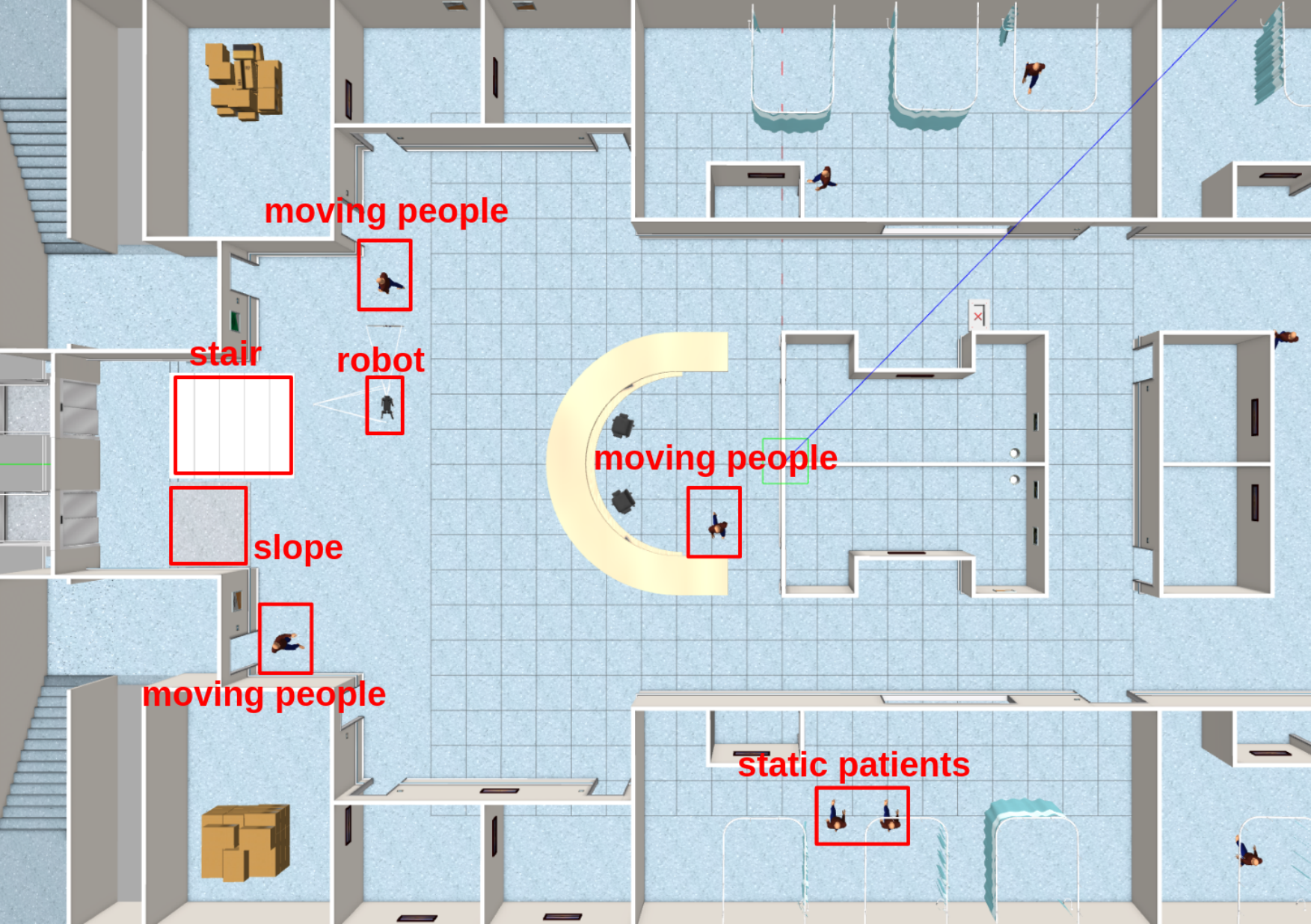}
	\caption{ The bird view of the second hospital simulation environment.}
	\label{bird view}
\end{figure}

\subsubsection{Reactive Control}
\begin{figure}[t]
	\centering
    \includegraphics[width=0.45\textwidth]{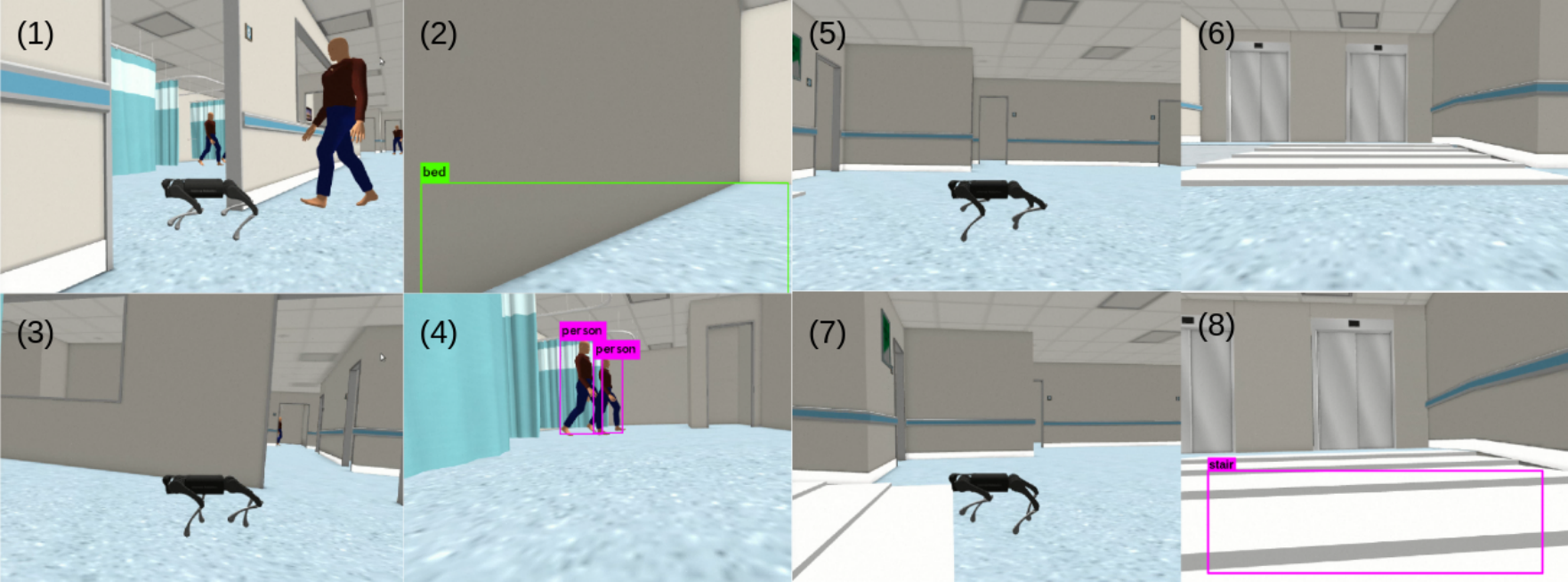}
	\caption{ Gazebo visualization of the reactive control when encountering environment changes.}
	\label{gazebo stair and ward}
	\end{figure}

In this section, we introduce a scene-understanding guided reactive motion control policy. To demonstrate the reactive control module, as shown in Fig. \ref{bird view}, a bigger and more complicated hospital environment is created. The quadruped robot needs to switch gaits to accommodate different terrains while performing high-level navigation tasks. We consider two terrains, the flat ground and stairs, in the hospital environment, which can be detected by YOLO-V3 implemented with MindSpore. On the flat ground, the quadruped robot uses the trot gait for motion efficiency. If stairs are encountered, the gait is switched to walk to ensure torso stability when climbing up the stairs or going down the stairs. The patient status is also considered when determining the gait of the quadruped robot since in the hospital scene the robot needs to carry out tasks simultaneously without disturbing patients at rest. As shwon in Fig. \ref{gazebo stair and ward}, if a static (e.g., sleeping) patient is detected in the ward, the quadruped robot will move quietly by switching to the walk gait to avoid the noise induced by the switching between the flight phase and the stance phase when trotting. If multiple moving people are detected, the quadruped robot will switch back to the trot gait for better motion. 


\subsection{Experiment \label{Experiment}}

 In the experiment, the constrained task is $\phi_c = \diamondsuit (water \land \Circle (\diamondsuit (bag \land \Circle (\diamondsuit human ^ {Room_B \vee Room_A})))) \vee \diamondsuit (bag \land \Circle (\diamondsuit (water \land \Circle (\diamondsuit human^{Room_B \vee Room_A}))))$, where $water$ represents the water bottle, $bag$ represents the school bag and $human$ represents the person. The task $\phi_c$ requires the quadruped robot to first find the water bottle and school bag and then hand them to the human, who might be in either area $A$ or area $B$. Similarly, $\phi_c$ is first translated to an LTL task specification $\phi = \diamondsuit (water \land \Circle (\diamondsuit (bag \land \Circle (\diamondsuit human)))) \vee \diamondsuit (bag \land \Circle (\diamondsuit (water \land \Circle (\diamondsuit human))))$ for off-line pre-planning. As shown in Fig. \ref{real_exp}(c), since the target objects are out of the robot's field of view, neither the school bag nor or the water bottle can be observed by the robot at the initial position 1. The robot goes to area $C$ and area $D$ sequentially according to the pre-planned trajectory. Unfortunately, none of the target objects can be observed and thus the gate mechanism determines that the subtask cannot be completed. Hence, the robot needs to explore the environment to find a way to complete the subtask. Since no position constraints are associated with the water bottle and the school bag and no relevant information available in its prior knowledge, the greedy mechanism is activated to guide the robot to complete the next subtask. Later, at the corner the robot finds the school bag and water bottle in area $A$ and the human in area $B$. The robot then goes to area $A$ to complete the task of packing the school bag and water bottle. After the gate mechanism confirms that two subtasks have been finished, the robot goes to area $B$ to hand the water and the school bag to the human.  More details are referred to the experiment video\footnote{\url{https://youtu.be/EE5MKiAuezc}}.

\section{CONCLUSIONS}
	
This work presents a vision-based reactive motion planning and control framework in unstructured dynamic environments. The effectiveness of our approach has been demonstrated via numerical simulation and physical experiments. Future work will be extended to heterogeneous multi-robot systems for more challenging collaborative tasks.

	\addtolength{\textheight}{-12cm}   
	



	\bibliographystyle{IEEEtran}
	\bibliography{reference}

\begin{thebibliography}{10}
\providecommand{\url}[1]{#1}
\csname url@samestyle\endcsname
\providecommand{\newblock}{\relax}
\providecommand{\bibinfo}[2]{#2}
\providecommand{\BIBentrySTDinterwordspacing}{\spaceskip=0pt\relax}
\providecommand{\BIBentryALTinterwordstretchfactor}{4}
\providecommand{\BIBentryALTinterwordspacing}{\spaceskip=\fontdimen2\font plus
\BIBentryALTinterwordstretchfactor\fontdimen3\font minus
  \fontdimen4\font\relax}
\providecommand{\BIBforeignlanguage}[2]{{%
\expandafter\ifx\csname l@#1\endcsname\relax
\typeout{** WARNING: IEEEtran.bst: No hyphenation pattern has been}%
\typeout{** loaded for the language `#1'. Using the pattern for}%
\typeout{** the default language instead.}%
\else
\language=\csname l@#1\endcsname
\fi
#2}}
\providecommand{\BIBdecl}{\relax}
\BIBdecl

\bibitem{bledt2018cheetah}
G.~Bledt, M.~J. Powell, B.~Katz, J.~Di~Carlo, P.~M. Wensing, and S.~Kim, ``Mit
  cheetah 3: Design and control of a robust, dynamic quadruped robot,'' in
  \emph{2018 IEEE/RSJ International Conference on Intelligent Robots and
  Systems (IROS)}.\hskip 1em plus 0.5em minus 0.4em\relax IEEE, 2018, pp.
  2245--2252.

\bibitem{fankhauser2018anymal}
P.~Fankhauser and M.~Hutter, ``Anymal: a unique quadruped robot conquering
  harsh environments,'' \emph{Research Features}, no. 126, pp. 54--57, 2018.

\bibitem{leslie2022robots}
M.~Leslie, ``Robots tackle darpa underground challenge,'' pp. 2--4, 2022.

\bibitem{li2018neural}
Y.~Li, R.~Cui, Z.~Li, and D.~Xu, ``Neural network approximation based
  near-optimal motion planning with kinodynamic constraints using rrt,''
  \emph{IEEE Transactions on Industrial Electronics}, vol.~65, no.~11, pp.
  8718--8729, 2018.

\bibitem{Cai2021b}
M.~Cai, M.~Hasanbeig, S.~Xiao, A.~Abate, and Z.~Kan, ``Modular deep
  reinforcement learning for continuous motion planning with temporal logic,''
  \emph{IEEE Robot. Autom. Lett.}, vol.~6, no.~4, pp. 7973--7980, 2021.

\bibitem{cai2023overcoming}
M.~Cai, E.~Aasi, C.~Belta, and C.-I. Vasile, ``Overcoming exploration: Deep
  reinforcement learning for continuous control in cluttered environments from
  temporal logic specifications,'' \emph{IEEE Robotics and Automation Letters},
  vol.~8, no.~4, pp. 2158--2165, 2023.

\bibitem{chen2020temporal}
G.~Chen, M.~Liu, and Z.~Kong, ``Temporal-logic-based semantic fault diagnosis
  with time-series data from industrial internet of things,'' \emph{IEEE Trans.
  Ind. Electron}, vol.~68, no.~5, pp. 4393--4403, 2020.

\bibitem{zhou2022multiple}
X.~Zhou, T.~Yang, Y.~Zou, S.~Li, and H.~Fang, ``Multiple subformulae
  cooperative control for multiagent systems under conflicting signal temporal
  logic tasks,'' \emph{IEEE Transactions on Industrial Electronics}, vol.~70,
  no.~9, pp. 9357--9367, 2022.

\bibitem{chen2011formal}
Y.~Chen, X.~C. Ding, A.~Stefanescu, and C.~Belta, ``Formal approach to the
  deployment of distributed robotic teams,'' \emph{IEEE Transactions on
  Robotics}, vol.~28, no.~1, pp. 158--171, 2011.

\bibitem{Cai2023}
M.~Cai, S.~Xiao, Z.~Li, and Z.~Kan, ``Optimal probabilistic motion planning
  with potential infeasible ltl constraints,'' \emph{IEEE Trans. Autom.
  Control}, vol.~68, no.~1, pp. 301--316, 2023.

\bibitem{li2021asymmetric}
Z.~Li, G.~Li, X.~Wu, Z.~Kan, H.~Su, and Y.~Liu, ``Asymmetric cooperation
  control of dual-arm exoskeletons using human collaborative manipulation
  models,'' \emph{IEEE Transactions on Cybernetics}, vol.~52, no.~11, pp.
  12\,126--12\,139, 2021.

\bibitem{li2019human}
Z.~Li, C.~Deng, and K.~Zhao, ``Human-cooperative control of a wearable walking
  exoskeleton for enhancing climbing stair activities,'' \emph{IEEE
  Transactions on Industrial Electronics}, vol.~67, no.~4, pp. 3086--3095,
  2019.

\bibitem{Vasile2020}
C.~I. Vasile, X.~Li, and C.~Belta, ``Reactive sampling-based path planning with
  temporal logic specifications,'' \emph{Int. J. Robot. Res.}, p.
  0278364920918919, 2020.

\bibitem{li2021safe}
Y.~Li, E.~M. Shahrivar, and J.~Liu, ``Safe linear temporal logic motion
  planning in dynamic environments,'' in \emph{2021 IEEE/RSJ International
  Conference on Intelligent Robots and Systems (IROS)}.\hskip 1em plus 0.5em
  minus 0.4em\relax IEEE, 2021, pp. 9818--9825.

\bibitem{cai2020learning}
M.~Cai, H.~Peng, Z.~Li, and Z.~Kan, ``Learning-based probabilistic ltl motion
  planning with environment and motion uncertainties,'' \emph{IEEE Transactions
  on Automatic Control}, vol.~66, no.~5, pp. 2386--2392, 2020.

\bibitem{li2022human}
Z.~Li, X.~Li, Q.~Li, H.~Su, Z.~Kan, and W.~He, ``Human-in-the-loop control of
  soft exosuits using impedance learning on different terrains,'' \emph{IEEE
  Transactions on Robotics}, vol.~38, no.~5, pp. 2979--2993, 2022.

\bibitem{li2020human}
Z.~Li, K.~Zhao, L.~Zhang, X.~Wu, T.~Zhang, Q.~Li, X.~Li, and C.-Y. Su,
  ``Human-in-the-loop control of a wearable lower limb exoskeleton for stable
  dynamic walking,'' \emph{IEEE/ASME transactions on mechatronics}, vol.~26,
  no.~5, pp. 2700--2711, 2020.

\bibitem{li2023human}
Z.~Li, Q.~Li, P.~Huang, H.~Xia, and G.~Li, ``Human-in-the-loop adaptive control
  of a soft exo-suit with actuator dynamics and ankle impedance adaptation,''
  \emph{IEEE Transactions on Cybernetics}, 2023.

\bibitem{Li2022}
Z.~Li, M.~Cai, S.~Xiao, and Z.~Kan, ``Online motion planning with soft metric
  interval temporal logic in unknown dynamic environment,'' \emph{IEEE Control
  Syst. Lett.}, vol.~6, pp. 2293--2298, 2022.

\bibitem{otte2016rrtx}
M.~Otte and E.~Frazzoli, ``Rrtx: Asymptotically optimal single-query
  sampling-based motion planning with quick replanning,'' \emph{Int. J. Robot.
  Res.}, vol.~35, no.~7, pp. 797--822, 2016.

\bibitem{kim2019highly}
D.~Kim, J.~Di~Carlo, B.~Katz, G.~Bledt, and S.~Kim, ``Highly dynamic quadruped
  locomotion via whole-body impulse control and model predictive control,''
  \emph{arXiv preprint arXiv:1909.06586}, 2019.

\bibitem{ding2021representation}
Y.~Ding, A.~Pandala, C.~Li, Y.-H. Shin, and H.-W. Park, ``Representation-free
  model predictive control for dynamic motions in quadrupeds,'' \emph{IEEE
  Transactions on Robotics}, vol.~37, no.~4, pp. 1154--1171, 2021.

\bibitem{baier2008}
C.~Baier and J.-P. Katoen, \emph{Principles of model checking}.\hskip 1em plus
  0.5em minus 0.4em\relax MIT press, 2008.

\bibitem{luo2021abstraction}
X.~Luo, Y.~Kantaros, and M.~M. Zavlanos, ``An abstraction-free method for
  multirobot temporal logic optimal control synthesis,'' \emph{IEEE Trans.
  Rob.}, vol.~37, no.~5, pp. 1487--1507, 2021.

\bibitem{bochkovskiy2020yolov4}
A.~Bochkovskiy, C.-Y. Wang, and H.-Y.~M. Liao, ``Yolov4: Optimal speed and
  accuracy of object detection,'' 2020.

\bibitem{ester1996density}
M.~Ester, H.-P. Kriegel, J.~Sander, X.~Xu \emph{et~al.}, ``A density-based
  algorithm for discovering clusters in large spatial databases with noise,''
  in \emph{kdd}, vol.~96, no.~34, 1996, pp. 226--231.

\bibitem{liao2022walking}
Q.~Liao, Z.~Li, A.~Thirugnanam, J.~Zeng, and K.~Sreenath, ``Walking in narrow
  spaces: Safety-critical locomotion control for quadrupedal robots with
  duality-based optimization,'' \emph{arXiv preprint arXiv:2212.14199}, 2022.

\bibitem{sleiman2021unified}
J.-P. Sleiman, F.~Farshidian, M.~V. Minniti, and M.~Hutter, ``A unified mpc
  framework for whole-body dynamic locomotion and manipulation,'' \emph{IEEE
  Robotics and Automation Letters}, vol.~6, no.~3, pp. 4688--4695, 2021.

\bibitem{bellicoso2016perception}
C.~D. Bellicoso, C.~Gehring, J.~Hwangbo, P.~Fankhauser, and M.~Hutter,
  ``Perception-less terrain adaptation through whole body control and
  hierarchical optimization,'' in \emph{2016 IEEE-RAS 16th International
  Conference on Humanoid Robots (Humanoids)}.\hskip 1em plus 0.5em minus
  0.4em\relax IEEE, 2016, pp. 558--564.

\end{thebibliography}

\end{document}